# LOCAL TRANSLATION SERVICES FOR NEGLECTED LANGUAGES

David Noever[1], Josh Kalin[2], Matthew Ciolino[1], Dom Hambrick[1], and Gerry Dozier[2]

[1]PeopleTec, Inc., 4901 Corporate Drive. NW, Huntsville, AL, USA
`david.noever@peopletec.com`

[2]Department of Computer Science and Software Engineering, Auburn University, Auburn, AL, USA
`jzk098@auburn.edu`

## ABSTRACT

*Taking advantage of computationally lightweight, but high-quality translators prompt consideration of new applications that address neglected languages. For projects with protected or personal data, translators for less popular or low-resource languages require specific compliance checks before posting to a public translation API. In these cases, locally run translators can render reasonable, cost-effective solutions if done with an army of offline, small-scale pair translators. Like handling a specialist's dialect, this research illustrates translating two historically interesting, but obfuscated languages: 1) hacker-speak ("l33t") and 2) reverse (or "mirror") writing as practiced by Leonardo da Vinci. The work generalizes a deep learning architecture to translatable variants of hacker-speak with lite, medium, and hard vocabularies. The original contribution highlights a fluent translator of hacker-speak in under 50 megabytes and demonstrates a companion text generator for augmenting future datasets with greater than a million bilingual sentence pairs. A primary motivation stems from the need to understand and archive the evolution of the international computer community, one that continuously enhances their talent for speaking openly but in hidden contexts. This training of bilingual sentences supports deep learning models using a long short-term memory, recurrent neural network (LSTM-RNN). It extends previous work demonstrating an English-to-foreign translation service built from as little as 10,000 bilingual sentence pairs. This work further solves the equivalent translation problem in twenty-six additional (non-obfuscated) languages and rank orders those models and their proficiency quantitatively with Italian as the most successful and Mandarin Chinese as the most challenging. For neglected languages, the method prototypes novel services for smaller niche translations such as Kabyle (Algerian dialect) which covers between 5-7 million speakers but one which for most enterprise translators, has not yet reached development. One anticipates the extension of this approach to other important dialects, such as translating technical (medical or legal) jargon and processing health records or handling many of the dialects collected from specialized domains (mixed languages like "Spanglish", acronym-laden Twitter feeds, or urban slang).*

## 1. INTRODUCTION

Technical advances in natural language models have propelled the democratization of foreign translation services. Given only bilingual sentence pairs, anyone can now write a lightweight, high-quality translator that learns from fewer than 20,000 examples. While text translation between multiple languages has formerly stood as a challenge to even large software enterprises [1], deep learning models (in fewer than one hundred lines of additional code) now routinely deliver translations that rival human experts. To translate languages that are too obscure (dialects) or intentionally obfuscated (computer-centric hacker-speak, "Leet", or "l33t" [2-3]), we take advantage of this low barrier-to-entry and build multiple novel translation services. To demonstrate how these translators work, we rely heavily on the neural network model [4] first demonstrated for English-German language pairs [5], one which as illustrated in Figures 1 and 2, used a long short-term memory (LSTM), recurrent neural network (RNN) model [6-8]. The successful translator thus should generalize to previously unseen phrases and word combinations without memorizing a given set of training sequences [4,9]. If the translator is computationally lightweight, it may find a useful niche to handle obscure or otherwise previously unapproachable translations [10-14] that document owners cannot disseminate into larger enterprise or cloud-hosted applications. We demonstrate these micro-models for multiple non-traditional translation tasks.

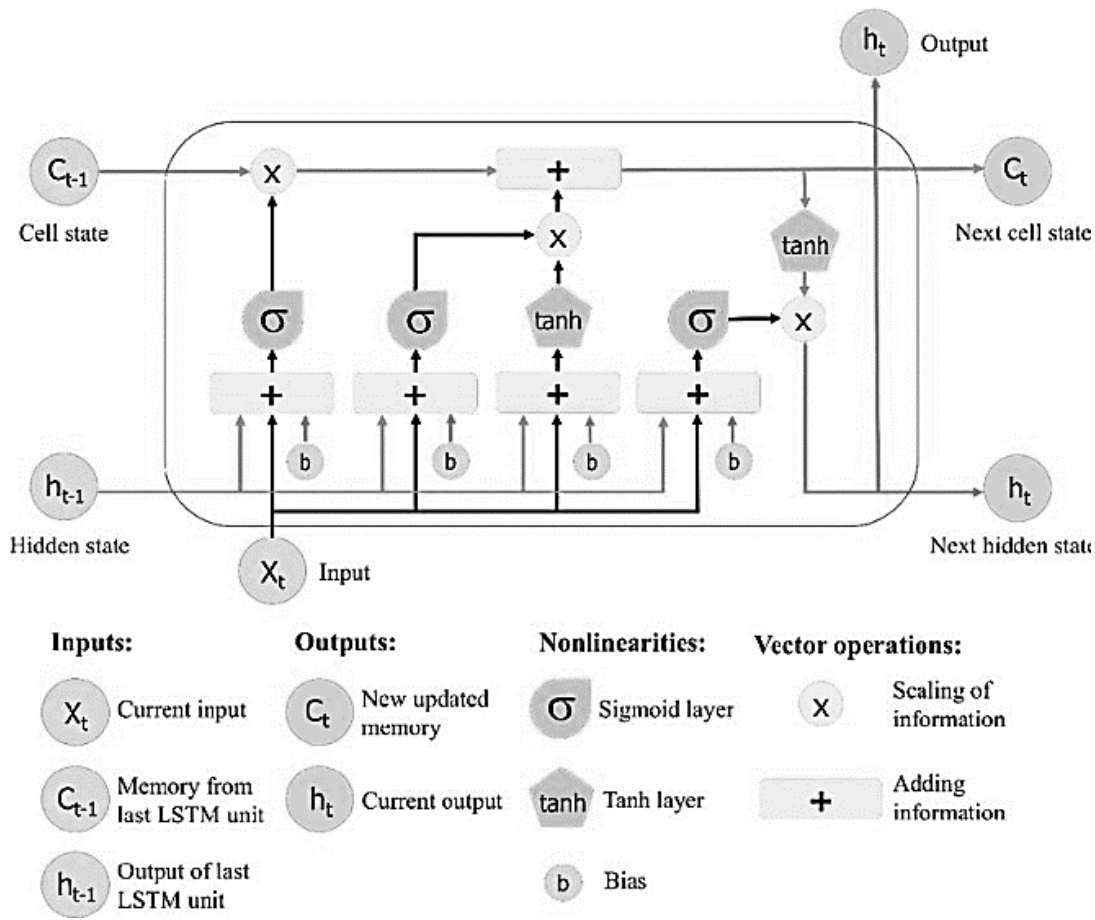

Figure 1. Long Short-Term Memory (LSTM) Block Diagram Illustrating the Information Flow and Weights to Solve Through Iterative Presentation of Training Data

To build initial confidence, we reproduce the main published findings for the local English-German translation service [4] and report the popular quantitative metric, BLEU, or the Bilingual Evaluation Understudy score [15]. This word-error rate provides a decimal score between 0 and 1, with a perfect match as one. The BLEU score compares the expected and predicted phrase(s) based on word selection and ordering while penalizing wrong substitutions either at a sentence- or corpus-level over an entire document.

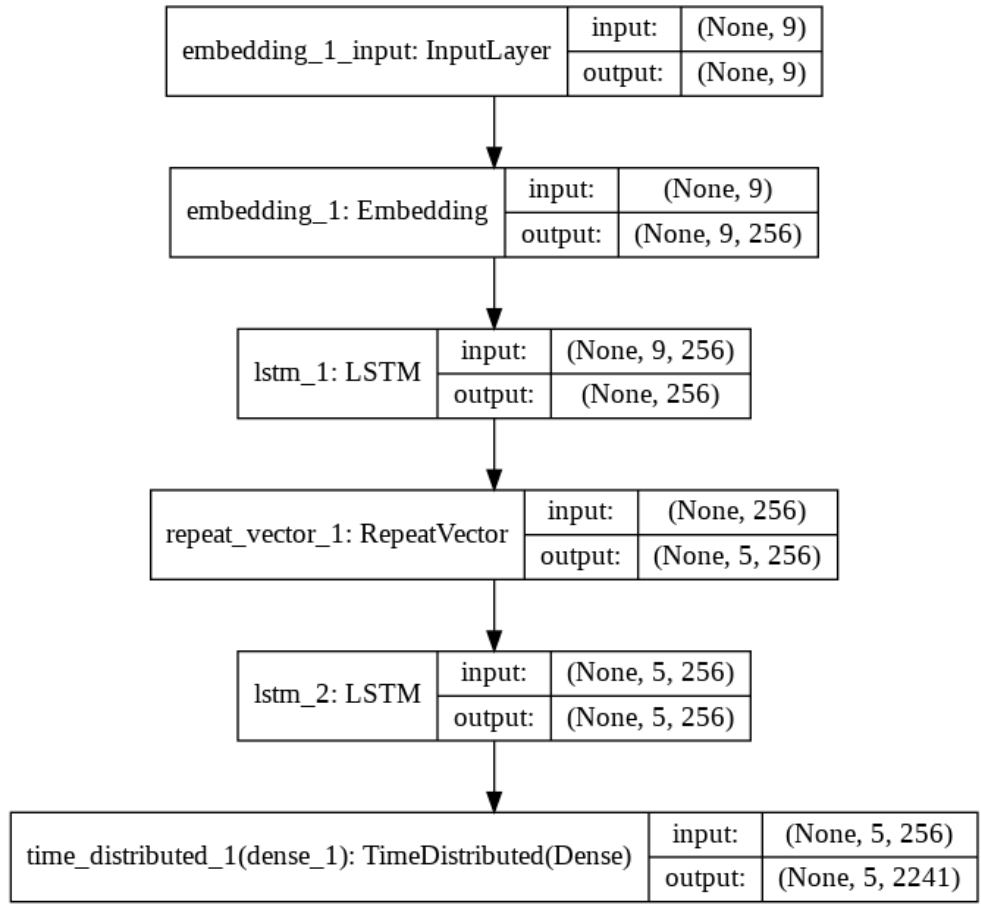

Figure 2. Long Short-Term Memory (LSTM) Recurrent Neural Network for Translation with Embedding Parameters for Each Stage.

As shown in Table 1, we subsequently generalize the LSTM to translate twenty-six other language pairs using the same basic LSTM approach with bilingual pairs: English-to-Hungarian, English-to-Turkish, English-to-Persian, etc. Each specific task depends heavily on training data consisting of 10,000 or more common phrase pairings [5,16-17], where the parallel input provides the first example in English and a second example translated into the target language. To get the translation contextually correct, this bilingual (source-target) or paired approach to language modelling applies no specific syntax rules or tags no parts-of-speech. In other words, the data presentation itself follows the classic "train-by-example" only, without any feature engineering or expert domain knowledge [18]. For this case, one can call the model "end-to-end", meaning it needs no human intervention to define the various verb tenses or to identify the past participles unique to any language pairing [19]. Given this minimal initiation step, the approach scales quickly to multiple instances of the translation challenge across different languages, dialects, and technical specializations (like medicine [20], law [21], finance [22], or in the present case, computer hobbyists [1-3]).

In contrast with state-of-the-art (SOTA) translators using much bigger (but more costly) training approaches [23-26], these standard translation tasks and their corresponding BLEU scores give a comparison for both language difficulty and fidelity [27]. It is worth emphasizing that these benchmarks provide comparative indications; the intention is not to compete or overtake much larger (but less wieldy) enterprise providers. The one interest in localizing these micro-models hinges therefore on handling self-contained or potentially confidential document translations that would not qualify for sharing as cloud-based solutions, but which could effectively need such local translators for many less popular languages [28-29]. Among these low-resource language translators, one obvious example involves translating health records [20] or other protected data [21-22] that may require specific compliance checks before posting to a public API, but which could render reasonable, cost-effective solutions if done with an army of local, small-scale models. A secondary use-case involves more edge-computing, where a lightweight or portable translator might enable contextual understanding of slang (e.g. informal regionalization, prison patter [29], "da Vinci-like mirror writing" [30], hacker-speak [1-3] or texting short-hand of popular acronyms like "LOL" and "IRL").

Table 1. Example Translation Pairs from English for the Same Input Phrase "It Was A Bomb". Training data consist of greater than 10,000 pairs from (or to) English from another language.

| **Language** | **Bilingual Sentence Pair** | **Language** | **Bilingual Sentence Pair** |
|---|---|---|---|
| English | It was a bomb | Spanish | fue una bomba |
| Mirror Writing | bmob a saw ti | Hungarian | Bomba volt |
| Leet Lite | 17 waz a b0mb | Russian | это была бомба |
| Leet Mid | 3y37 vv45 4 80448 | Chinese | 那是一颗炸弹 |
| Leet Hard | ai1 JLaye5 aye 6ohem6 | Ukrainian | це була бомба |
| Turkish | O bir bombaydı | Hebrew | זו הייתה פצצה |
| Italian | Era una bomba | Dutch | Het was een bom |
| Italian-to-Eng | It was a bomb | German | Es war eine Bombe |
| Kabyle | yebbeεzeq | Arabic | كانت قنبلة |
| Berber | Tella tleffuɣt | Persian | این یک بمب بود |
| Korean | 그것은 폭탄이었다 | Romanian | era o bombă |
| French | C'était une bombe | Portuguese | Foi uma bomba |
| Japanese | 爆弾でした | Marathi | तो एक बॉम्ब होता |
| Finnish | se oli pommi | Polish | to była bomba |
| Bulgarian | беше бомба | Serbian | беше бомба |
| Czech | byla to bomba | Greek | ήταν μια βόμβα |

One specialty language that extends beyond existing enterprise solutions has received attention because of its novelty and obfuscation challenges: hacker-speak. Often called l33t, this written dialect adopted by computer enthusiasts since the 1990's originally evolved to discuss uncensored topics while avoiding simple language filtering on bulletin boards and hacker forums. As shown in Figure 3, Leet-speak uses an alternative alphabet of numbers and symbols to replace various letters in words [1-3]. As an innovative language strategy, Leet might be one of the first adversarial attacks on machine-driven filters [31].

Interesting examples of this Leet letter-substitution pattern exist mainly as a one-way generator, where a given English phrase can spawn multiple degrees of obfuscated variants. For example, given an English sentence like "I like computers", one completely deterministic interpreter would programmatically substitute some common single letters: English "I" becomes number "1", "A" becomes number "4", etc. Most authors 1-3,32] of this type of one-way generator conclude that the inverse problem of universally rendering English from Leet offers considerably more challenges given the "many-to-one" difficulty of reversing what might appear as random alphabetic substitutions.

```
A => 4   @    /-\  /\   ^    aye ə   ci  λ   z
B => 8   |3   6    13   |3   ß   ]3
C => (   <    ¢    {    ©    sea see
D => |)  [)   ∂    ])   I>   |>  0   ð   cl
E => 3   £    &    €    [-   ə
F => |=  ]=   }    ph   (=   ſ
G => 6   9    &    (_+  C-   gee jee (γ, cj
H => |-| #    ]-[  [-]  )-(  (-) :-:     }{  }-{ aych
I => !   1    |    eye  3y3  ai  ¡
J => _|  _/   ]    ¿    </   _)  ĵ
K => X   |<   |X   |{   ƙ
L => 1   7    |_   £    |    |_  lJ  ¬
M => 44  /\/\      |\/|      em  |v| IYI IVI [V] ^^  nn //\\//\\      (V)
N => |\| /\/  //\\//           И     [\] <\> {\} //  ₪   []\[]    ]\[  ~
O => 0   ()   oh   []   ¤    Ω
P => |*  |o   |°   |>   |"   ?   9   []D |7  q   þ   ¶   □   |D
Q => 0_  0,   (,)  <|   cue  9   ¶
R => |2  2    /2   I2   |^   |~  1z  ®   |2  [z  |`  12  Я   .-  ʁ
S => 5   $    z    §    es
T => 7   +    -|-  1    '][' †
U => |_|      (_)  Y3W  M    µ   [_] \_/ \_\ /_/
V => \/  √    \\//
W => \/\/     vv   '//  \\'  \^/ (n) \X/ \|/ \_|_/     \\//\\//     \_:_/
X => %   ><   Ж    }{   ecks ×   *   )(  ex
Y => j   `/   `(   -/   '/   Ψ   φ   λ   Ч   ¥
Z => 2   ≥    ~/_  %    3    7_
```

Figure 3. Substitution Dictionary for Hacker-Speak. The left-side of the arrow (=>) shows English character and right-side options for substitutions as numbers, letters, or symbols.

To bound the problem, the present work adopts a three-tiered approach to what might more simply span a scale of easy to hard. As summarized in Table 1 and Figure 3, the hacker dialect splits into three categories for "lite, mid (or medium), and hard" ranges, depending on the depth and variation of single letter substitutions.

For concreteness in Figure 3, the lite-variant substitutes the first column, mainly numbers, for only "{s,e,i,o,t}" with case-sensitivities (e.g., "Shoot me" becomes "Zh007 m3"). The medium-variant, or mid, substitutes for all letters with mostly numbers or phonetics (e.g. "Shoot me" becomes "5aych007 443"). Because the generation of such dialects is deterministic in the forward direction (English-to-Leet), we want to probe this language modeling in the reverse case (Leet-to-English). The hard-variant substitutes with a random replacement chosen from all available columns, thus transforming the translation into a more interesting, non-deterministic "many-to-one" pattern recognition problem (e.g. "Shoot me" becomes "5aychoh07 em3" and "5aych0oh1 nn3", etc.). Like most languages, the harder "Leet" variants can give several possible outputs for the same input sentence depending on the depth and quantity of single letter substitutions [32]. When given a highly obfuscated text example in Leet, how would one automate the generation of plausible English interpretations? Finally, given the paucity of good training examples as language pairs, we want to generate three novel data sets of more than 100,000 phrase pairs that can seed more advanced approaches in machine learning and translation training.

To make this final goal more realizable, we train a Generative Pre-Trained Transformer (GPT-2, [33]) to generate variant text in full pairings for English-Leet while preserving the required tab-separated format for future modeling. Remarkably, this GPT-2 approach can capture both the substance and style for a host of fascinating language tasks [34] including preserving the syntax and page spacing required for poetry, rhyme, music lyrics, film scripts, or plays [35]—all just from mimicking a training dataset (and document layout) while fine-tuning the last decoder step [36] of a massively pre-trained example curated from internet-scale text samples.

Previous work has demonstrated that LSTM models offer a reasonable micro-translator compared to the often-gigantic transformers that dominate current Natural Language Understanding (NLU) approaches. A published figure of merit estimates that the training cost for a typical transformer model exceeds $3 million in computer time [25]. An alternative metric compares the 100+ million parameters for most SOTA models like the Bidirectional Encoder Representation from Transformers (BERT) [26] to the typical 2-10 million parameters to achieve high-quality translations using the current LSTM [4].

The main original contributions of this work include 1) generation of the first public English-to-Leet language pairs that can seed large scale, deep learning models with millions of potential training examples; 2) quantifiable comparisons for dialects like hacker-speak when translated and contrasted against other complex languages like Hungarian, German, Dutch and Turkish; 3) a text generation model that provides almost limitless novel candidates for automated expansion of the full English-Leet dictionary. We believe given the many global languages (7,111) [29], their spoken dialects, and uncountable pairings available for study, this approach of micro-modeling makes variations more amenable to practical analysis. One outcome of this work therefore features an automated translation model to render (for example) entire Shakespearean plays (with original English content) but written in the "foreign" language of Leet.

## 2. METHODS

This research applies an LSTM network to language pairs in multiple examples ranging from Hungarian to hacker-speak. Figures 1 and 2 graphically shows the LSTM model in schematics with multiple layers, word embeddings, and its more than 2.66 million tuneable parameters. The bilingual pairs [16] vary in length, but an average of 20,000 phrases includes about 110,000 words or the average scale of a 300-page novel as training data.

### 2.1. Dataset Preparation

Bilingual sentence pairs provide the core training data for Turkish, Hungarian, Dutch, German, and all other models. More than 81 phrase pairs of varying length [16] have been published with examples from the Tatoeba Project [5], as part of the Anki language flashcards software [17]. The input for training is tab-delimited bilingual sentence pairs, with a third column attributing the reference for the (human) translated cases. The largest parallel training example for English sentences totals over 150,000 phrases and pairs that prompt input with Turkish output for one corpus. In this crowd-sourced collection, the smallest corpus registers fewer than 100 bilingual sentences (as English-to-Bosnian), which provides little guidance for actual translation services. Since one experiment builds small models for low-resource languages such as Kabyle (Algerian dialect), we include those outputs as original contributions to the translators' capabilities. In the absence of published bilingual pairs, we supplemented example phrases using the Google Translation API and Google Sheets [23-24] to generate representative pairs for Korean, Persian, and Romanian.

To generate three levels of Leet or hacker-speak, we begin with the 150,000 English-only sentences clipped from the largest corpus [16] and deterministically substitute for the alphabet in a one-for-one (lite), many-for-one (mid or medium), or random (hard) dictionary approach. To extract a literal rough translation, this naïve approach might compare to a character-based substitution pattern that transliterates Greek, Cyrillic, and Latin alphabets. In our case, the origins and evolution of Leet allow the generation of three large datasets with enough complexity and repeatability that the hacker-speak offers an experimental candidate for exploring dialects. Figure 3 summarizes the three tiers investigated and illustrates their substitution rules with examples in Table 1. We load these bilingual pairs as a look-up dictionary to train the LSTM sequences as a recurrent neural network with encoder stages and translated outputs.

## 2.2. Model Parameters

*LSTM layers and tuning.* The LSTM builds long-term associations or language context in a manageable set of neuron-like gates by over-weighting essential connections (like the closely related semantic terms, "lion-cat") in a process called "attention" [7]. Tuned for the key task of learning to translate, the LSTM slowly forgets over time ("short-term memory") [8]. As illustrated schematically in Figure 1, this "forgetful", recurrent network architecture in part learns to regulate information flow (and weights) while modeling ordered sequences. Another motivation for the LSTM architecture stems from the ability of its "input, output, and forget gates" to learn continuously without vanishing gradients [8] and thus avoid learning plateaus found in standard recurrent networks.

*GPT-2 Transformer for text generation.* As a final task, we augment the Leet training set using text generation of bilingual pairs. By fine-tuning GPT-2 [33-36], we train a text generator on Shakespeare's plays, but with the hard Leet pair in a second tab-separated column that supports further translator training cycles. This method applies the python libraries for gpt-2-simple [36] to customize the last layer of a pre-trained decoder on a style and content previously not part of the transformers larger training cycle [33]. The standard GPT-2 tuning parameters include batch size of 5, temperature of 0.7, text length = 1023 (maximum).

## 2.3. Quantitative Metrics

We use the well-known [15, 27] Bilingual Evaluation Understudy Score (BLEU) to measure the approximate similarity between reference translations by humans and the machine-translated text. For a reference and test phrase, the BLEU constructed n-grams (n=1-4) and counts the matches independent of word position [4]. Higher scores indicate more overlap and better candidate translations.

## 3. RESULTS

Tables 2-4 show that the LSTM recurrent neural networks translate complex bilingual pairs with lightweight models, including examples for English to the 26 traditional languages in order of easier to harder as quantified by the BLEU score. Table 2 shows the translations for one example phrase "she just left" in the obfuscated language pairs (English-Leet) and mirror writing, then summarizes the BLEU scores for word-error rates over the whole test data set previously not seen by the LSTM (approximately 2,000 phrases). Tables 3-4 summarize the translation success using the same method, but for both traditional and low-resource language pairs. To approximate the language difficulty to English, Table 3 shows the rank-order best performing models with a BLEU score greater than "High-Quality" while Table 4 shows the more challenging language pairs with scores "Good" or less.

The LSTM furthermore works symmetrically by swapping columns and Table 3 shows the English-to-Italian vs. Italian-to-English models work as a level between "expert" to "fluent" in reverse inputs. The success interpretation derives from the closeness of the translator's output on unseen test data compared to a set of high-quality reference translations [27]. Since the goal of these relatively lighter weight models is not state-of-the-art (but rather passable extensions to previously unmodeled languages), the results of Table 2 achieve an initial range between understandable to better-than-human translations.

A match in BLEU-4 implies the 4-gram (e.g. 4-word order) is preserved between reference and candidate but represents the cumulative scores from 1 to n with a weighted geometric mean. This score penalizes missed word substitutions and ordering up to 4 tokens in a sentence sequence compared to the reference translation. The qualitative conclusion of whether a model achieved fluency derives from literature [27] scales that bin values mainly by 0.1 intervals ranging from useless (<0.1), hard to get the gist (0.10-0.19), the gist is clear but with grammatical errors (0.20-0.29), understandable to good translation (0.30-0.40), high quality (0.40-0.50), fluent (0.50-0.60), and quality often better than humans (> 0.60).

Table 2. Bilingual Evaluation Understudy (BLEU) scores for Translator Per Bilingual Pairs in the Obfuscated Examples

| Language | Example Translation | BLEU-1 | BLEU-2 | BLEU-3 | BLEU-4 | Interpretation |
|---|---|---|---|---|---|---|
| English starter | She just left | - | - | - | - | - |
| Mirror Writing | tfel tsuj ehS | 0.95 | 0.93 | 0.90 | 0.72 | May Exceed Human |
| Leet Lite | Zh3 juz7 l3f7 | 0.63 | 0.55 | 0.46 | 0.22 | Fluent |
| Leet Mid | 5aych3 _|M57 13ph7 | 0.37 | 0.25 | 0.14 | 0.02 | Understandable |
| Leet Hard | esaych3 JY3Wes1 lJ3ph7 | 0.23 | 0.11 | 0.26 | 0.32 | Gist is clear |

Table 3. Bilingual Evaluation Understudy (BLEU) scores for Translator Per Bilingual Pairs

| Foreign Language | Example Translation | BLEU-1 | BLEU-2 | BLEU-3 | BLEU-4 | Interpretation |
|---|---|---|---|---|---|---|
| English starter | She just left | - | - | - | - | - |
| Italian | se n'è appena andata | 0.79 | 0.72 | 0.66 | 0.45 | May Exceed Human |
| Google API Italian | ha appena lasciato | 0.74 | 0.64 | 0.59 | 0.49 | May Exceed Human |
| Portuguese | ela acabou de sair | 0.64 | 0.52 | 0.47 | 0.31 | May Exceed Human |
| French | elle vient juste de partir | 0.60 | 0.49 | 0.43 | 0.28 | Fluent |
| Dutch | Ze is net vertrokken | 0.59 | 0.46 | 0.40 | 0.28 | Fluent |
| Spanish | ella se acaba de ir | 0.54 | 0.41 | 0.35 | 0.22 | Fluent |
| Finnish | hän juuri lähti | 0.54 | 0.41 | 0.36 | 0.25 | Fluent |
| Google API Russian | она только что ушла | 0.53 | 0.42 | 0.38 | 0.28 | Fluent |
| German | Sie ist gerade gegangen | 0.53 | 0.40 | 0.31 | 0.12 | Fluent |
| Turkish | O henüz gitti | 0.53 | 0.40 | 0.33 | 0.18 | Fluent |
| Italian-to-English | She just left | 0.50 | 0.36 | 0.26 | 0.12 | Fluent |
| Marathi | ती नुकतीच निघून गेली | 0.46 | 0.36 | 0.32 | 0.21 | High-Quality |
| Hungarian | Éppen kiment | 0.45 | 0.31 | 0.25 | 0.14 | High-Quality |
| Romanian | ea tocmai a plecat | 0.40 | 0.25 | 0.19 | 0.10 | High-Quality |

For further comparison, in June 2020, Google Translate reports a peak in BLEU score around 0.53 and an average score between 0.2-0.3 across a distribution of its 108 languages translated into English [23]. To score the LSTM against Google Translate, we took a common set of the human-translated Italian and Russian and compared the larger Google models for relative BLEU scores with our LSTM results. For Italian, the LSTM (BLEU-1=0.79) compares well with the Google API (0.74). For Russian, the LSTM (BLEU-1=0.38) performs less well compared to Google API (0.53) on the same inputs.

Table 4. Bilingual Evaluation Understudy (BLEU) scores for Translator Per Bilingual Pairs

| Foreign Language | Example Translation | BLEU-1 | BLEU-2 | BLEU-3 | BLEU-4 | Interpretation |
|---|---|---|---|---|---|---|
| Greek | μόλις έφυγε | 0.38 | 0.27 | 0.22 | 0.13 | Good |
| Russian | Она ушли | 0.38 | 0.27 | 0.20 | 0.04 | Good |
| Persian | او فقط رفت | 0.37 | 0.21 | 0.16 | 0.08 | Good |
| Czech | právě odešla | 0.37 | 0.23 | 0.17 | 0.09 | Good |
| Ukrainian | Вона щойно пішла | 0.36 | 0.25 | 0.19 | 0.07 | Good |
| Polish | Ona właśnie wyszła | 0.29 | 0.17 | 0.13 | 0.06 | Gist is clear |
| Serbian | управо је отишла | 0.28 | 0.19 | 0.16 | 0.09 | Gist is clear |
| Hebrew | היא בדיוק עזבה | 0.26 | 0.16 | 0.11 | 0.04 | Gist is clear |
| Bulgarian | тя просто си тръгна | 0.25 | 0.15 | 0.12 | 0.06 | Gist is clear |
| Korean | 그녀는 방금 떠났다 | 0.26 | 0.11 | 0.06 | 0.02 | Gist is clear |
| Berber | Imir-a kan ay tedda | 0.22 | 0.14 | 0.09 | 0.13 | Gist is clear |
| Kabyle | ad yakan | 0.19 | 0.12 | 0.10 | 0.04 | Hard to get gist |
| Arabic | غادرت للتو | 0.15 | 0.06 | 0.04 | 0.01 | Hard to get gist |
| Japanese | 彼女はちょうど去った | 0.06 | 0.01 | 0.06 | 0.09 | Hard to get gist |
| Chinese | 她走了 | 0.04 | 0.13 | 0.21 | 0.24 | Hard to get gist |

From the original 2002 BLEU proposal [15], an example of human translators working on 500 sentences reported BLEU=0.3468 against four references and 0.2571 against two references. For better language models [37], the following rank-order minima criteria measured acceptance criteria across the 1-to-n-gram version of BLEU scores: BLEU-1: 0.401 to 0.578, BLEU-2: 0.176 to 0.390, BLEU-3: 0.099 to 0.260, and BLEU-4: 0.059 to 0.170. When sorted by translation proficiency in Figure 4, Romanian is the cut-off for an acceptable ("High Quality") translator of non-obfuscated languages, and Leet-mid to Leet-hard offer challenges at this standard.

While BLEU provides a simple approximate measure of translated fidelity, its shortcomings include word order sensitivities that may appear awkward or out-of-place to a native speaker ("translation-ese"). By example [4], changing no sentence meaning with "quick" for "Fast" and "lazy" for "sleep" in a test sentence can change a baseline score of 0.75 ("fluent") to 0.48 ("good"), but offers no upgrade in contextual meaning to a human interpreter. An example is a token substitution from a dictionary that similarly might drop the Spanish term for "house" as "casa" arbitrarily in a sentence to get the gist of its meaning but would prove tedious to a reader of several hundred pages of this poor translation-ese. To account for this bias, we tested the machine models with human scorers for confirmation. Two volunteers scored 100 randomly selected German-to-English translation pairs to validate the BLEU scores against knowledgeable human expectations. One volunteer did the same task for Italian-to-English and Spanish-to-English translated outputs. The native speakers of German and Italian scored the machine-translations lower on average than the BLEU metric. For the 100 randomly selected German-to-English pairs, the humans rated the system output between "Gist is clear" and "Gist with errors". For Italian, the humans rated the system output between "Good" and "Gist with errors". Given 5 rankings (0.1="Useless"; 0.2="Gist"; 0.3="Gist with errors"; 0.4="Good"; 0.5="Fluent"), the average German score was 0.234 with a variation between scores (0.224-0.244). This result compares to the automated similarity score for the overall "fluent" rated German (BLEU-1=0.53). The Italian rated higher with the human translator as average (0.38), compared to the overall "expert" rated Italian (BLEU-1=0.79). The final Spanish score from a (non-native) student volunteer followed a similar trend with a "Gist" rating (0.215) vs. calculated (BLEU-1=0.54). Overall the human translators scored the BLEU rating approximately half the machine-rated BLEU-1 score, but consistent with the word-ordered BLEU-3 and 4 scores as one might expect for contextual meanings stripped of the jumbled "translation-ese" of machine versions.

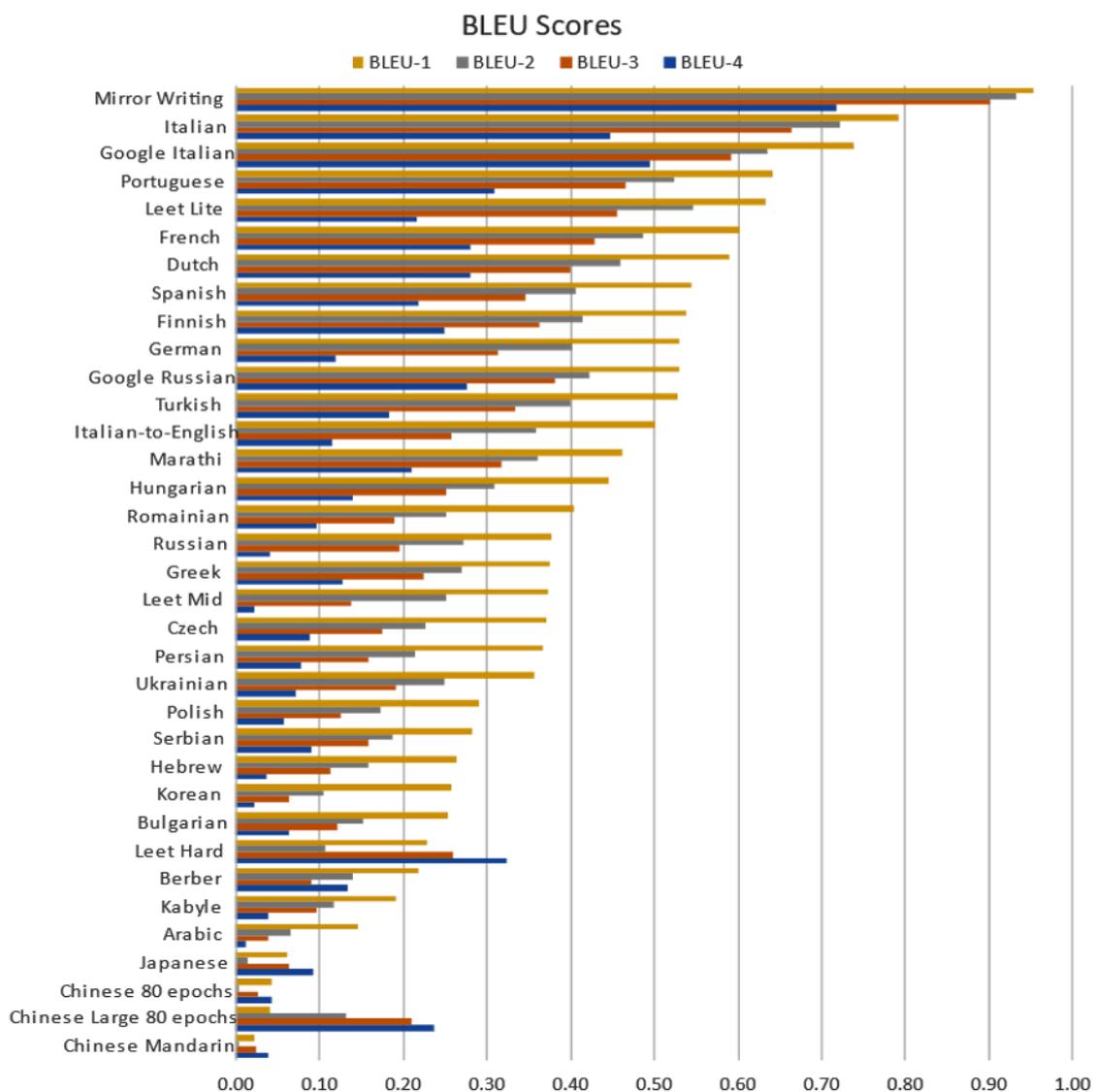

Figure 4. BLEU scores for tested languages from 1-4 n-grams.

Obfuscated languages represent a second challenge for these micro-models and benefit from the automated similarity scoring against expectations, given that no human speakers of these languages exist. For da Vinci's mirror writing [30] illustrated in Figure 5, the LSTM captures the pattern generation algorithm with near-perfect (BLEU-1=0.95) equivalency between actual and predicted from expert translation (BLEU-1=1). This outcome reflects the more deterministic character for the substitution pattern and given 2.6 million or more model parameters and only 6000 unique tokenized words, would prove disappointing if not this repetitively pattern-matching. It is worth noting that Leonardo [30] employed mirror (cursive) handwriting, not block type fonts, so his stylized diaries would not show the same understandability to an untrained eye as implied by these models. After all, the cumbersome nature of writing backward offered a way to hide controversial or private opinions from powerful censors or authoritarian critics. Other evidence [30] may support the more conventional explanation for mirror-writing obfuscation: as a left-handed writer, right-to-left scripting reduced the chance of ink smear as Leonardo wrote quickly.

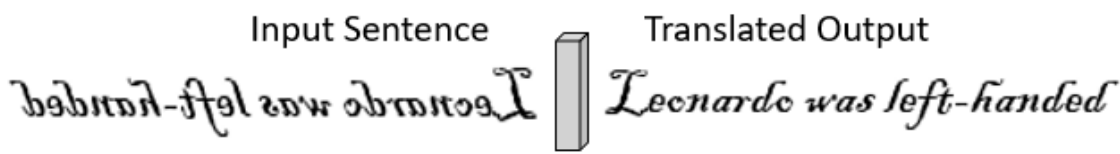

Figure 5. Long short-term memory (LSTM) recurrent neural network serves as the symbolic "mirror" to translate reversed writing as obfuscated text

This overall translation development proves simple enough to generalize the LSTM models to a subset of obfuscated languages such as da Vinci mirror writing or Leet-speak evolved from the hacker communities. In the latter cases, the simpler ("lite") number for letter substations compare favorably (BLEU-1=0.63) with the non-obfuscated pairs with traditional languages like German (BLEU-1=0.52) and Dutch (BLEU-1=0.59). This equivalency implies high-quality translations. For the deeper ("mid") and more randomized ("hard") substitution Leet-speak variants, the BLEU-1 scores decline to a moderately good translation. A typical value greater than 0.15 offers an opportunity for post-editing corrections to achieve publishable translations, readable but not without needing additional interpretation as shown in Figure 5 for the difficult Persian and Korean examples. While most BLEU scores improve with adding larger numbers of examples, the relatively constant limit of 35,000 training phrases here provides a baseline to evaluate the difficulty of each bilingual variant as a fair experimental comparison.

```
['where did you learn french'  '어디 프랑스어를 배웠어요']
['we go fishing once in a while'  '우리는 가끔 낚시를 가서']
["have you read the owner's manual"  '당신이 사용 설명서 읽게']
['is that scientifically proven'  '즉 과학적으로 입증']
["i didn't know where to put the package"  '패키지를 넣어 어디에서 몰랐다']
["i'd like to change my reservation"  '난 내 예약을 변경하려면']
["it's still light out"  '여전히 빛을 밖으로이다']
['he's now short of money'  'او در حال حاضر با کمبود پول']
['tom has low blood pressure'  'تام دارای فشار خون پایین']
["things aren't as bad as they seem"  'همه چیز به عنوان بد نیست به نظر می رسد']
['add more water'  'آب بیشتری اضافه کنید']
['tom sat with his legs crossed'  'تام نشسته با پاهای خود عبور']
['both tom and mary work as models'  'هر دو مدل تام و کار مریم به عنوان']
['she spends a lot of money on shoes'  'او صرف مقدار زیادی از پول در کفش']
```

Figure 6. Example output test translation sequences in Korean (top) and Persian (bottom).

The overall translation production pipeline now demonstrates the classic boot-strapping method of augmenting a smaller amount of data (<10,000 bilingual pairs), training a translator to map between any two languages, then customizing a larger text generator to replenish the starting step with a virtually unlimited quantity of more high-quality inputs in the correct format for continuous learning (>100,000). The additional output of the text generator includes GPT-2's remarkable mimicry of both substance and style [33-36], such that entire original plays can be rendered in translated output and returned as inputs to additional model refinements. Put simply, one can write a vast corpus in Leet from Shakespearean English, as illustrated in Figure 7.

| Generated Shakespeare | Companion Leet Paired at Generation Time |
| --- | --- |
| A trickle of blood is the malady of man! | 4 5trN31 0ph 8100clj 3y35 7aych3 44480j 0ph 444N, |
| Heavens, am I a beast, | aych0vv 3y3 7M2N'cl, |
| How I turn'd, as a stud now king his aunties: | 45 4 57Mcl N0vv X3y3N6 aych3y35 4MNcl33y335: |
| And stony reptile in his breast | 4Ncl 50N7j 7234222 3y3N aych3y35 823457 |
| Oil'd, for sure, liberal, and cruel degree. | 0seeM7'cl, ph02 50, 1372M3, 7aych3y36aychN7j, 4Ncl see2M3. |
| Hang in holding (to hold it more awfully) | aych4N6 3y3N aych3423y3N6 (70 aych01cl 3y37 44023aych4130vv32j) |
| I do bleed and bleed through drops. | 3y3 cl0 8133cl 4Ncl 8100cl 7aych23y37aych32 cl37715 cl2M5. |
| O, say th' most deadly lesson! Certain unhappiness will | 0, 54j 7aych' 44057 cl34cl3y3 13cl131372j MN3y36aych423j vv3y311 |
| Be rife against hire. All this disparity | 83 2M932 44j aych443. 411 7aych3y35 cl35see3947023 |
| The natural causes of amity. The meek shall err | 7aych3 N47M23 see4M253 ph4see77025. 7aych3 443X3cl 531V3 |
| Last do end with a shivering brow. | 1451 0oh 3Ncl vvai7aych aye esaych3y3V3l21Njee 13lzohJL |

Figure 7. GPT-2 results for co-generating original Shakespeare plays simultaneously with the tab-separated and line-for-line Leet translation pair as bilingual inputs for bootstrapping subsequent training cycles

## 3. CONCLUSIONS

To translate the estimated 7,111 unique languages [29], the combinatorial explosion of training data needed for bilingual sentence phrases exceeds known storage and computational limits. Handling the 10 most used languages alone requires 945 unique bilingual pairs for training sets and translation models [38]. In February 2020, Google's Translation API supported 5 new languages covering about 75 million (previously neglected) speakers, but also raising their total translatable language count to 108 and thus greater than $10^{86}$ bilingual pairs [24]. This combinatorial challenge motivates much of the current work by others [39] to build universal or multilingual translators [1] rather than smaller bilingual pairs [4].

This research highlights the potential for LSTM architectures to capture enough sequence order to build lightweight translators using only bilingual phrase pairs (<20,000 pairs). Using this approach, the work compares three types of alternative translation problems: 1) low-resource or neglected languages like Kabyle; 2) many popular languages where certain privacy restrictions may govern their transmission to public-API services; and 3) slang or dialect specialist domains like Leet, mirror-writing, or technical jargon.

For the novel translator of hacker-speak, the models perform best when one-for-one substitutions ("lite") govern the pair generator and decline in BLEU for one-to-many possible candidates ("mid" and "hard"). We show near perfect performance of de-obfuscating mirror-writing.

We show the LSTM architecture compares favorably for some popular languages like Italian and outperforms large public translation services. Extending these models for many high-ASCII character languages like Russian, we capture the gist in an English translation. We demonstrate that the bilingual pairs work in both directions, with English to Italian and vice versa. It is worth noting that the LSTM works reversibly with the bilingual pairs and requires no significant architectural modification to learn Russian-to-English for example versus English-to-Russian. We finally build large-scale generators for text pairs that provide future inputs to LSTM translators. As expected for data-dependent, deep learning approaches, larger translation dataset [40] offer the greatest opportunities for improving future translators. Using the transformer architecture for encoding-decoding multilingual text sources [39] offers a powerful but much larger modelling effort, one which works best for public APIs and not the local micro-models as this research demonstrates. For advancing the present methods for the difficult languages to tokenize like Chinese, recent work [41] using raw byte-pair encoders offers promising methods to continue developing sequence-to-sequence models.

## ACKNOWLEDGMENTS

The authors would like to thank the PeopleTec Technical Fellows program for encouragement and project assistance.

**Authors**

David Noever, Ph.D. has research experience with NASA and the Department of Defense in machine learning and data mining. Dr. Noever has published over 100 conference papers and refereed journal publications. He earned his Ph.D. in Theoretical Physics from Oxford University, as a Rhodes Scholar.

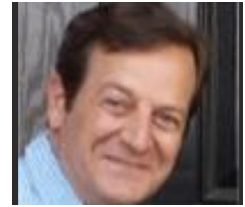

Matt Ciolino has research experience in deep learning and computer vision. He received his Bachelor's in Mechanical Engineering from Lehigh University. Matt is pursuing graduate study in computer vision and machine learning at Georgia Tech.

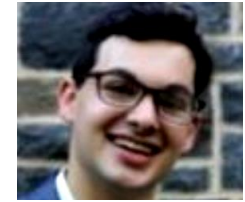

Josh Kalin is a physicist and data scientist focused on the intersections of robotics, data science, and machine learning. Josh received his Bachelor's in Physics, and Masters' in Mechanical Engineering from Iowa State University, and Computer Science from Georgia Tech.

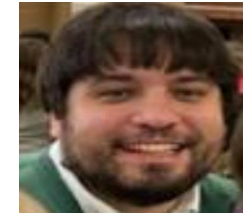

Dom Hambrick has 7 years of experience as a research engineer at Georgia Tech Research Institute. Dom received his Bachelor's in Physics with a minor in Mathematics from the University of Alabama, Huntsville. He has pursued graduate study in computer science and mathematics,

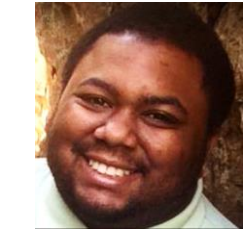

Gerry Dozier, Ph.D., is the Charles D. McCrary Eminent Chair Professor in the Department of Computer Science & Software Engineering at Auburn University. Dr. Dozier is the director of the Artificial Intelligence and Identity Research (AI2R) Lab. Dr. Dozier has published over 140 conference and journal publications. He earned his Ph.D. in Computer Science from North Carolina State University.

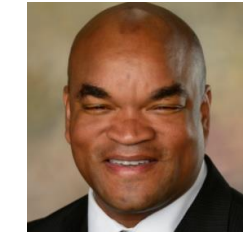